\documentclass[aps,prl,preprint,superscriptaddress]{revtex4-2}
\usepackage{amsmath,amssymb,bm}
\usepackage[colorlinks=true,linkcolor=blue,citecolor=blue,urlcolor=blue]{hyperref}

\usepackage{amsmath,amssymb,amsfonts}
\usepackage{geometry}
\usepackage{graphicx}
\usepackage{booktabs}
\usepackage{multirow}

\begin{document}

\title{Contextual Control without Memory Growth in a Context-Switching Task}

\author{Song-Ju Kim}
\email{kim@sobin.org}
\affiliation{SOBIN Institute LLC, Kawanishi, Hyogo, Japan}

\date{\today}

\begin{abstract}
Context-dependent sequential decision making is commonly addressed either by providing context explicitly as an input or by increasing recurrent memory so that contextual information can be represented internally. We study a third alternative: realizing contextual dependence by intervening on a shared recurrent latent state, without enlarging recurrent dimensionality. To this end, we introduce an intervention-based recurrent architecture in which a recurrent core first constructs a shared pre-intervention latent state, and context then acts through an additive, context-indexed operator.

We evaluate this idea on a context-switching sequential decision task under partial observability. We compare three model families: a label-assisted baseline with direct context access, a memory baseline with enlarged recurrent state, and the proposed intervention model, which uses no direct context input to the recurrent core and no memory growth. On the main benchmark, the intervention model performs strongly without additional recurrent dimensions.

We also evaluate the models using the conditional mutual information \(I(C;O \mid S)\) as a theorem-motivated operational probe of contextual dependence at fixed latent state. For task-relevant phase-1 outcomes, the intervention model exhibits positive conditional contextual information. Together, these results suggest that intervention on a shared recurrent state provides a viable alternative to recurrent memory growth for contextual control in this setting.
\end{abstract}

\maketitle

\section{Introduction}
\label{sec:introduction}

Context-dependent behavior is a basic requirement for sequential decision making under partial observability.
Such problems are commonly formalized as partially observable Markov decision processes (POMDPs), and recurrent neural networks are a standard way to integrate information over time when the current observation is not sufficient on its own \cite{KaelblingLittmanCassandra1998,HochreiterSchmidhuber1997,HausknechtStone2015,Bakker2001RLLSTM,KapturowskiEtAl2019R2D2,IglEtAl2018DVRL}.
In many sequential tasks, the same local observation must be interpreted differently depending on a latent task condition, a phase variable, or an external cue.
In such settings, the central challenge is not merely to remember past observations, but to implement conditional reuse of internal state: the agent must act on a shared recurrent representation while still allowing behavior to switch according to context.
This emphasis on how state is represented is also related to earlier work on predictive state representations \cite{LittmanSuttonSingh2001PSR}.

A standard way to handle contextual dependence is to provide the context explicitly as an additional input.
When this is possible, the model can condition its behavior directly on the label.
Another common strategy is to increase recurrent capacity so that contextual information can be represented internally.
Both approaches are useful baselines, but neither directly addresses the architectural question that motivates this paper:
\emph{can contextual dependence be implemented without enlarging recurrent memory and without concatenating the context token to the recurrent-state update?}

This question is related to recent information-theoretic analyses of contextuality under single-state representations \cite{Kim2026DecisionDynamics,Kim2026Obstruction,Kim2026NoGo,Spekkens2005,AbramskyBrandenburger2011}.
In particular, Kim \cite{Kim2026NoGo} studies classical single-state ontological models in which a fixed ontic state space is reused across interventions.
Under that constraint, Theorem~1 states that if observable statistics remain context dependent, then any such model requires an auxiliary contextual variable satisfying
\[
H(M_{\mathrm{aux}})\ge I(C;O\mid \lambda) > 0,
\]
where \(C\) denotes the intervention or context, \(\lambda\) the shared ontic state, \(O\) the observable outcome, and \(M_{\mathrm{aux}}\) an auxiliary contextual variable \cite{Kim2026NoGo}.
Here we write the theorem's auxiliary variable as \(M_{\mathrm{aux}}\), rather than \(M\), to avoid confusion with our memory baseline \(M\).
The key point is that the obstruction is not simply one of state-space size.
Rather, it arises from the requirement to reuse a common representational substrate across contexts while still expressing different context-dependent outcome statistics.
Rather than treating the theorem as a literal model of our recurrent agents, we use it as a motivating perspective on minimal contextual resources under shared-state reuse.

Our setting is not a literal ontological-model test of that theorem.
In the theorem, \(\lambda\) is an ontic variable in a classical probabilistic representation.
In our learning setup, the relevant conditioning variable is instead a learned recurrent latent state.
We therefore do \emph{not} claim a direct numerical verification of the theorem in full generality.
Instead, we use the theorem as a motivating resource-accounting picture and estimate an \emph{operational analogue}, namely \(I(C;O\mid S)\) at fixed recurrent latent state \(S\), to test whether task-relevant contextual effects remain in observable outcomes.

In this paper, we explore a concrete architectural alternative motivated by this perspective.
We introduce an \textbf{intervention-based recurrent model} in which a recurrent core first constructs a shared pre-intervention latent state, and context then acts through an additive, context-indexed operator:
\[
z_t' = z_t + \alpha D_{c_t}(z_t).
\]
This design differs from both direct label conditioning and memory expansion.
Instead of feeding context into the recurrent-state update or increasing recurrent dimensionality, the model keeps a shared latent representation and realizes contextual dependence by applying a context-specific transformation to that representation.
Our main claim is that this mechanism is sufficient to produce strong contextual control \emph{without memory growth} in the present benchmark.

To evaluate this idea, we study a sequential gridworld benchmark, \textbf{a context-switching sequential decision task}, in which the agent must solve a two-phase task with a single context switch inside each episode.
The same local observation can correspond to different target goals depending on the current phase and order condition.
This makes the benchmark a minimal but nontrivial testbed for context-dependent recurrent control.
Importantly, phase-1 reward and phase-1 success are counted only after phase-0 success, so the task specifically stresses whether a model can perform the required context-dependent reassignment of goals rather than exploit a degenerate shortcut.

We compare three model families:
(i) a \textbf{label-assisted baseline} L that directly observes the context token,
(ii) a \textbf{memory baseline} M that removes direct context input and instead enlarges recurrent hidden state,
and (iii) the proposed \textbf{intervention model} I that removes direct context input to the recurrent core but applies a context-indexed operator to a shared latent state.
This comparison isolates three distinct implementations of contextual dependence:
explicit conditioning, internal memory growth, and intervention on a shared recurrent representation.

Our empirical results show that the proposed intervention model performs strongly on the main context-switching benchmark.
It is competitive with the strongest memory-based baseline while requiring no increase in recurrent dimensionality.
We also examine the information-theoretic picture through the operational quantity \(I(C;O\mid S)\).
A key observation is that the outcome definition \(O\) matters:
when \(O\) is defined in a task-relevant way---for example, in terms of target-goal-related phase-1 outcomes---the intervention model exhibits positive conditional contextual information.
This supports the view that the model's contextual effect is expressed at the level of goal-directed outcomes rather than merely at the level of isolated primitive actions.

The main contributions of this paper are therefore threefold:
\begin{enumerate}
    \item We propose an \textbf{intervention-based recurrent architecture} that implements contextual dependence without enlarging recurrent memory.
    \item We introduce a controlled benchmark, \textbf{the context-switching task}, that isolates the architectural problem of context-dependent phase switching within a shared recurrent state.
    \item We provide both behavioral evidence and an information-theoretic operational probe showing that the proposed model realizes meaningful contextual control, while clarifying that this should be interpreted as an empirical analogue motivated by the single-state theorem rather than as a complete numerical verification of that theorem itself.
\end{enumerate}

Taken together, our results support a simple but important conclusion:
contextual dependence need not be implemented only by explicit label access or by enlarging recurrent memory.
A shared recurrent state, combined with a context-indexed intervention mechanism, can already provide a strong and interpretable solution in this setting.

More broadly, the same architectural question appears in larger agentic systems, where a shared internal state must support task-phase switching, role switching, or tool-selection control without relying only on ever larger memory.

\section{Methods}
\label{sec:methods}

\subsection{Problem setting}
\label{sec:problem_setting}

We study contextual control in a sequential gridworld task, which we refer to as \textbf{the context-switching task}.
Each episode is defined on a \(9\times 9\) maze with one start location and two candidate goals, denoted by \(G1\) and \(G2\).
At each time step, the agent receives a \textbf{local \(3\times 3\) observation} around its current position together with a \textbf{binary context token} \(c_t \in \{A,B\}\), and chooses one of four discrete actions,
\[
a_t \in \{\uparrow,\downarrow,\leftarrow,\rightarrow\}.
\]

A key property of the context-switching task is that the context changes \textbf{once per episode}.
Let \(t_{\mathrm{switch}}\) denote the switch time.
Before the switch, the episode is in \textbf{phase 0}; after the switch, it is in \textbf{phase 1}.
We consider two fixed-order settings:
\[
\text{AB}: \quad c_t = A \text{ in phase 0}, \quad c_t = B \text{ in phase 1},
\]
\[
\text{BA}: \quad c_t = B \text{ in phase 0}, \quad c_t = A \text{ in phase 1}.
\]

The target goal depends on the context.
In the AB setting, the target is \(G1\) in phase 0 and \(G2\) in phase 1; in the BA setting, the target is \(G2\) in phase 0 and \(G1\) in phase 1.
Thus, solving the task requires not only navigation, but also context-dependent switching of the target goal within a single episode.

\begin{figure}[t]
    \centering
    \includegraphics[width=\linewidth]{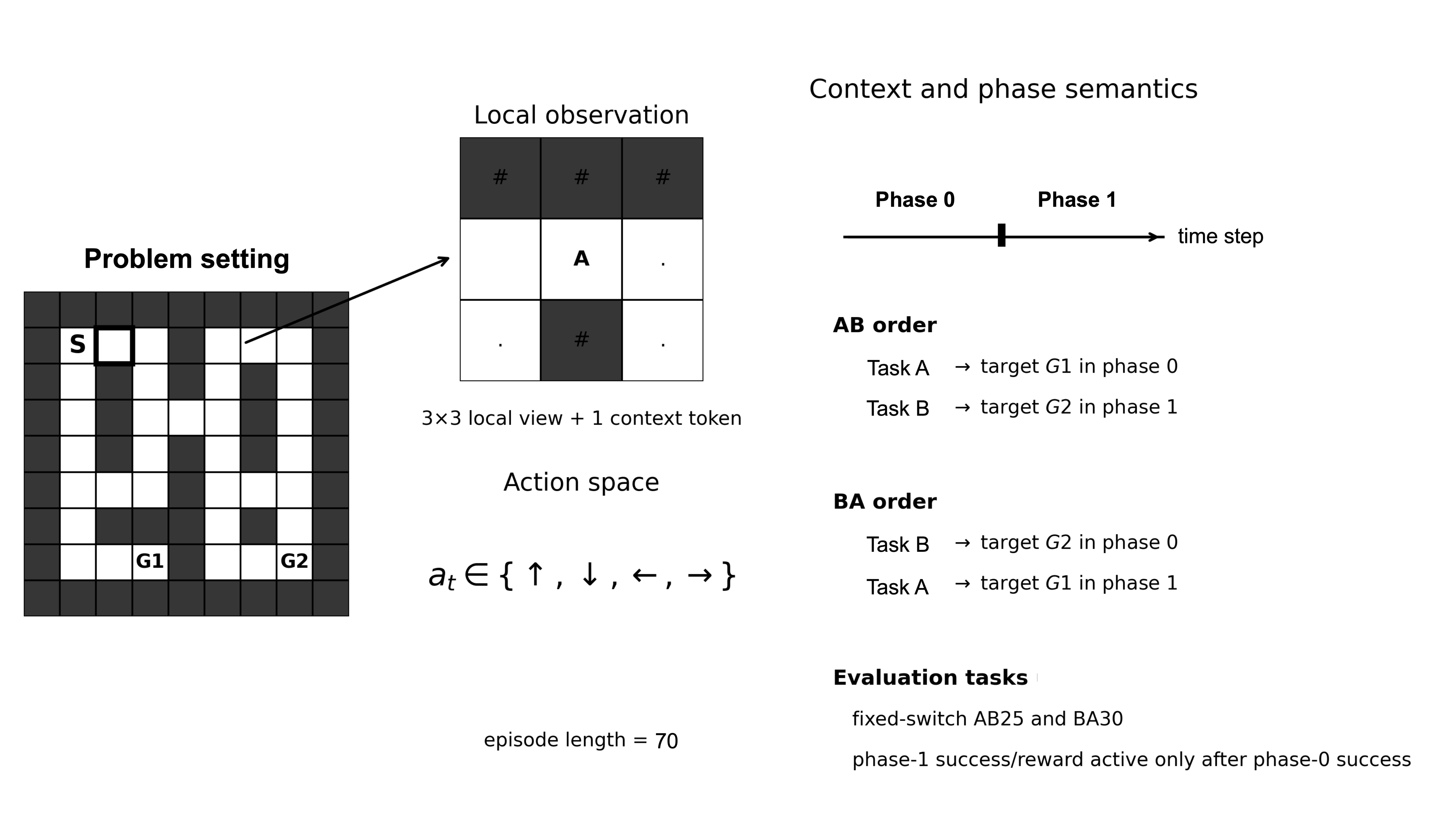}
    \caption{
    Problem setting of the context-switching sequential decision task.
    The agent acts in a \(9\times 9\) maze and observes a local \(3\times 3\) view together with a context token.
    The task contains two phases within each episode.
    In the AB order, the target is \(G1\) in phase 0 and \(G2\) in phase 1; in the BA order, the target is \(G2\) in phase 0 and \(G1\) in phase 1.
    The main benchmark conditions AB25 and BA30.
    Phase-1 success and reward are counted only after phase-0 success.
    }
    \label{fig:problem_setting}
\end{figure}

Figure~\ref{fig:problem_setting} summarizes the task.
The environment is a partially observable \(9\times 9\) maze with two candidate goals and a single within-episode context switch.
The agent receives a local \(3\times 3\) observation and must act according to the current phase-dependent target-goal mapping.
The main experiments focus on the conditions AB25 and BA30, with phase-1 success gated on prior phase-0 success.

Our main experiments focus on two conditions:
\[
\text{AB25} \quad \text{and} \quad \text{BA30},
\]
where the switch occurs at time step 25 or 30, respectively.
The context-switching task itself is not restricted to these two switch times.
In the main text, however, we focus on AB25 and BA30 as representative benchmark conditions, while broader switch-time robustness is left outside the main claims of this paper.
These representative conditions were chosen for the main benchmark comparison, rather than because the task formulation itself is restricted to those switch schedules.

An important design choice is that \textbf{phase-1 reward and phase-1 success are enabled only after phase-0 success}.
In other words, the agent cannot obtain full credit for phase 1 without first solving phase 0.
This phase-1-only gate removes degenerate strategies that ignore the first phase and directly optimize for the second phase.
As a result, differences between models are expected to appear primarily in \textbf{phase 1}, where context-dependent goal selection is most critical.

The reward includes a standard step cost and positive reward for reaching the correct goal, together with additional shaping and penalties for undesirable behaviors such as repeatedly hitting blocked cells or reaching the wrong goal.
However, the central difficulty of the context-switching task is not reward engineering per se; it is the need to implement \textbf{contextual dependence} under a recurrent representation that is shared across phases.

For clarity, throughout the paper we write the local observation as \(x_t\), the context token as \(c_t\), and the recurrent latent state as \(z_t\).
The learning problem is therefore to realize a policy
\[
\pi(a_t \mid x_{\le t}, c_{\le t})
\]
that switches appropriately between the two target-goal mappings within a single episode.

\subsection{Model definitions}
\label{sec:model_definitions}

We compare three recurrent model families, denoted by \textbf{L}, \textbf{M}, and \textbf{I}.
Their differences are summarized in Table~\ref{tab:model_definitions}.
All three models use the same LSTM-based recurrent backbone \cite{HochreiterSchmidhuber1997} and differ only in \textbf{how contextual information enters the computation}.

\paragraph{L: label-assisted recurrent baseline.}
The \textbf{L} model directly receives the context token as part of the observation.
Let \(\phi(\cdot)\) denote the feature extractor and \(h_{t-1}\) the recurrent hidden state.
Then the latent state is computed as
\[
z_t = \mathrm{LSTM}(\phi([x_t,c_t]), h_{t-1}).
\]
Thus, the context label is explicitly available at every step.
This model serves as an oracle-style reference baseline: it does not need to reconstruct the contextual variable from the recurrent dynamics alone, because contextual information is already present in the recurrent update.

\paragraph{M: memory-based baseline.}
The \textbf{M} model removes direct context input and instead increases recurrent capacity.
Its latent state is computed as
\[
z_t = \mathrm{LSTM}(\phi(x_t), h_{t-1}),
\qquad
\dim(z_t)=d+m,
\]
where \(d\) is the base recurrent size and \(m\) is the additional memory dimension.
In this model, the context token is \textbf{not} provided as a direct input to the recurrent core.
Any contextual dependence must therefore be represented \textbf{implicitly} in the enlarged recurrent state.
This model tests the standard strategy of solving contextual dependence by allocating additional internal memory.

\paragraph{I: intervention-based recurrent model.}
The \textbf{I} model is the main proposal of this paper.
As in M, the recurrent core does not directly receive the context token:
\[
z_t = \mathrm{LSTM}(\phi(x_t), h_{t-1}).
\]
However, instead of enlarging the recurrent state, I applies a \textbf{context-indexed intervention} to the shared latent state:
\[
z_t' = z_t + \alpha D_{c_t}(z_t),
\]
where \(D_{c_t}\) is a context-dependent linear operator and \(\alpha\) is a scalar intervention strength.
Concretely, in the binary-context case,
\[
z_t'=
\begin{cases}
z_t+\alpha D_A(z_t), & c_t=A,\\[4pt]
z_t+\alpha D_B(z_t), & c_t=B.
\end{cases}
\]
The modulated state \(z_t'\), rather than \(z_t\), is then passed to the policy and value heads.

In the implementation used in this paper, the intervention operators are learned bias-free linear maps on the latent space:
\[
D_A(z)=W_A z,\qquad D_B(z)=W_B z,
\]
where \(W_A,W_B\in\mathbb{R}^{d\times d}\) are trainable parameters.
The intervention is therefore implemented as
\[
z_t' = z_t + \alpha W_{c_t} z_t.
\]
Here \(\alpha\) is a fixed scalar hyperparameter, and the operator weights are initialized at zero so that the effective map starts near the identity at the beginning of training. We use a small fixed \(\alpha\) so that the intervention acts as a controlled residual perturbation of the shared latent state, while the zero initialization keeps the initial mapping close to the identity and avoids introducing a large context-specific distortion at the start of training. In the present setup, we found \(\alpha=0.1\) to work better than the larger values \(\alpha=0.2\) and \(\alpha=0.3\) that we also tested.

This intervention mechanism is conceptually related to condition-dependent modulation of intermediate representations \cite{deVriesEtAl2017CBN,PerezEtAl2018,DumoulinEtAl2018FeatureWise,KimSongBengio2017DLN}, but it is applied here to a recurrent latent state and is specifically designed to test whether contextual control can be implemented \emph{without} context concatenation in the recurrent-state update and \emph{without} recurrent memory growth.
The key idea is that the recurrent core builds a shared \textbf{pre-intervention latent state} \(z_t\), while contextual dependence is realized by an additive operator acting on that latent state.
In this way, the model can implement context-dependent behavior \textbf{without increasing recurrent dimensionality}.
This is the central architectural hypothesis evaluated in the paper.

\begin{figure}[t]
    \centering
    \includegraphics[width=\linewidth]{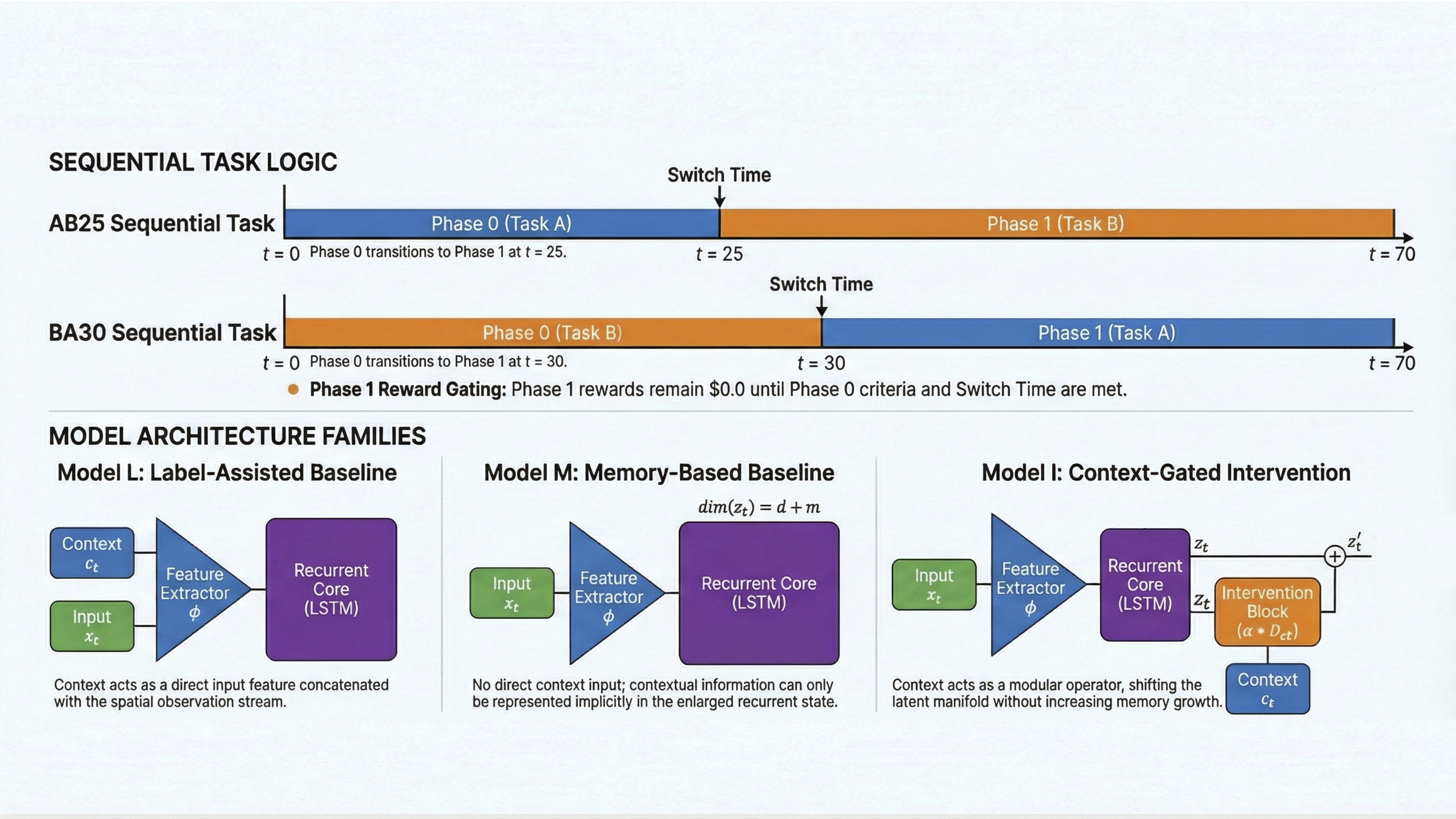}
    \caption{
      Overview of the benchmark conditions and model families.
The upper row provides a compact timeline view of the two representative benchmark conditions, AB25 and BA30.
The lower row compares the three model families.
    L directly receives the context token together with the spatial observation.
    M removes direct context input and instead enlarges the recurrent state from \(d\) to \(d+m\).
    I also removes direct context input from the recurrent core, but applies a context-indexed residual intervention to a shared pre-intervention latent state, yielding \(z_t' = z_t + \alpha D_{c_t}(z_t)\).
    This comparison isolates three distinct implementations of contextual dependence: explicit conditioning, memory expansion, and intervention on a shared pre-intervention latent state.
    }
    \label{fig:model_overview}
\end{figure}

Figure~\ref{fig:model_overview} provides a compact overview of the benchmark conditions and the three model families.
The upper row summarizes the representative benchmark timelines AB25 and BA30, while the lower row focuses on the architectural comparison.
The label-assisted baseline L uses direct context access, the memory baseline M allocates additional recurrent capacity, and the proposed intervention model I realizes contextual dependence by applying a context-indexed residual operator to a shared pre-intervention latent state.
This design makes it possible to test whether contextual control can be achieved without enlarging recurrent dimensionality.

\begin{table}[t]
\centering
\caption{Summary of the three model families. \(x_t\) denotes the local observation, \(c_t\) the context token, \(h_{t-1}\) the recurrent state, and \(m\) the extra memory size used only in M.}
\label{tab:model_definitions}
\begin{tabular}{llll}
\toprule
Model & Context access & Extra memory growth & Update rule / interpretation \\
\midrule
L & direct label input & no &
$z_t=\mathrm{LSTM}(\phi([x_t,c_t]), h_{t-1})$ \\
&&& oracle-style reference baseline \\
\addlinespace
M & no direct label & yes &
$z_t=\mathrm{LSTM}(\phi(x_t), h_{t-1}),\ \dim(z_t)=d+m$ \\
&&& memory baseline with enlarged hidden state \\
\addlinespace
I & operator only & no &
$z_t=\mathrm{LSTM}(\phi(x_t), h_{t-1}),\quad z_t' = z_t + \alpha D_{c_t}(z_t)$ \\
&&& intervention-based recurrent model \\
\bottomrule
\end{tabular}
\end{table}

Conceptually, the three models differ as follows:
\begin{itemize}
    \item \textbf{L}: context is available as an explicit input to the recurrent update;
    \item \textbf{M}: context must be represented implicitly in a larger recurrent memory;
    \item \textbf{I}: context acts as an operator on a shared recurrent latent state after the recurrent update.
\end{itemize}

This comparison isolates the main question of the paper: whether contextual dependence can be realized by \textbf{intervention on a shared state}, rather than by either direct label access or memory growth.

\subsection{Training and evaluation protocol}
\label{sec:training_eval}

All models are trained under a matched recurrent RL setup so that the comparison isolates the role of contextual access and memory growth.
The main experiments use the tasks AB25 and BA30 described above.
For each condition, we train \textbf{10 random seeds} for each model family.

The main training budget is \textbf{300k environment steps} per run.
Unless otherwise stated, we use the same base recurrent dimensionality \(d=32\) across models.
For the memory baseline, we evaluate multiple additional memory sizes,
\[
m \in \{8,16,32,64\}.
\]
For the intervention model, we use the same recurrent core size \(d\) as in L and apply a fixed intervention strength \(\alpha\) (set to \(0.1\) in the implementation used here). In preliminary tuning, we also tested \(\alpha=0.2\) and \(\alpha=0.3\), but \(\alpha=0.1\) gave the most satisfactory overall results in the present setup.
We therefore treat \(\alpha=0.1\) as a fixed design choice in this study, while leaving a more systematic sensitivity analysis over \(\alpha\) and its interaction with other hyperparameters to future work.

We report the following evaluation metrics:
\begin{enumerate}
    \item \textbf{Success-both}: whether the agent solves both phases within an episode.
    \item \textbf{Phase-0 success rate} and \textbf{phase-1 success rate}: the fraction of evaluation episodes in which each phase is solved.
    \item \textbf{Average return}: the mean episode return over evaluation episodes.
    \item \textbf{Wrong-goal statistics}: auxiliary diagnostics such as the rate of hitting the incorrect goal.
\end{enumerate}

Among these, the most important metrics for the present paper are \textbf{success-both} and \textbf{phase-1 success rate}.
Because phase-1 credit is gated on phase-0 success, failures in contextual control are expected to manifest primarily as reduced phase-1 performance.

\subsection{Information-theoretic quantities}
\label{sec:info_quantities}

A secondary goal of this paper is to connect the empirical behavior of the models to the information-theoretic theorem of Kim \cite{Kim2026NoGo}.
In that theorem, the central statement is formulated for classical single-state ontological models as
\[
H(M_{\mathrm{aux}}) \ge I(C;O \mid \lambda) > 0,
\]
where \(C\) is the intervention or context, \(\lambda\) is a shared ontic state, \(O\) is an observable outcome, and \(M_{\mathrm{aux}}\) is an auxiliary contextual variable not contained in \(\lambda\) \cite{Kim2026NoGo}.
We use the notation \(M_{\mathrm{aux}}\) here to avoid confusion with our memory baseline \(M\); the two are conceptually distinct.

We do not instantiate that theorem literally.
In our learning setup there is no ontological-model variable \(\lambda\) given a priori.
Instead, we define an \textbf{operational analogue} using the recurrent latent state and estimate
\[
I(C;O \mid S),
\]
where \(S\) denotes a model-dependent latent conditioning variable.
This quantity is used as an empirical probe of contextual dependence at fixed latent state in the context-switching benchmark.

In the present setting, we instantiate the variables as follows:
\begin{itemize}
    \item \(C\): the binary context variable (\(A\) or \(B\));
    \item \(S\): the latent state used for conditioning in the estimator;
    \item \(O\): a task-relevant observable outcome.
\end{itemize}

For the intervention model, the natural choice of \(S\) is the \textbf{pre-intervention latent state} \(z_t\), since the model is explicitly designed so that contextual influence enters only after this shared state is formed.
For the other models, we use the analogous recurrent latent state before the policy head when constructing the same estimator.
This does not make the three models ontologically identical, but it does provide a matched operational comparison.

\paragraph{Counterfactual estimator.}
To estimate contextual influence at fixed latent state, we use a counterfactual construction.
For each stored latent state \(s\), we compute the outcome distribution under both contexts:
\[
p_0(\cdot \mid s) = p(O \mid S=s, C=0),
\qquad
p_1(\cdot \mid s) = p(O \mid S=s, C=1).
\]
Given a context prior \(w_0,w_1\), we define the mixture
\[
m(\cdot \mid s)=w_0 p_0(\cdot \mid s)+w_1 p_1(\cdot \mid s),
\]
and estimate
\[
\widehat I(C;O\mid S)
=
\mathbb{E}_{s}
\left[
w_0 \,\mathrm{KL}\!\left(p_0(\cdot\mid s)\,\|\,m(\cdot\mid s)\right)
+
w_1 \,\mathrm{KL}\!\left(p_1(\cdot\mid s)\,\|\,m(\cdot\mid s)\right)
\right].
\]
Under a uniform context prior, \(w_0=w_1=\tfrac12\), this reduces to the Jensen--Shannon divergence between the two counterfactual outcome distributions, averaged over latent states.

For the intervention model, this counterfactual is implemented by keeping the same pre-intervention latent state fixed and changing only the intervention context, rather than recomputing the recurrent state from a modified observation.

\paragraph{Outcome definitions.}
A crucial methodological choice is the definition of \(O\).
In principle, one could define \(O\) as a one-step primitive action.
However, in the context-switching task the context-dependent effect is expressed primarily through \textbf{goal-directed behavior in phase 1}, rather than through a single primitive action in isolation.
We therefore consider \textbf{task-relevant outcome definitions} derived from the local geometry and the counterfactual action distributions.

The main outcome definitions used in the paper are:
\begin{itemize}
    \item \texttt{target\_hit}: whether the next step reaches the context-appropriate target goal;
    \item \texttt{goal3}: a three-way outcome distinguishing \(\{\text{other}, \text{target}, \text{wrong}\}\).
\end{itemize}

These quantities are computed from the local \(3\times 3\) observation and the counterfactual action distributions under the two contexts.
In the main text, we focus on \textbf{phase 1} and use a \textbf{uniform prior over contexts}, since this is the regime in which contextual goal selection is most directly expressed.
We also examined primitive-action outcomes in preliminary analyses, but they were less informative for the present benchmark and are therefore omitted from the main text.

We emphasize that our goal is \textbf{not} to claim a complete numerical verification of every assumption and term in the theorem of \cite{Kim2026NoGo}.
Rather, we use \(I(C;O\mid S)\) as a principled empirical probe motivated by the same single-state resource-accounting picture.


\section{Results}
\label{sec:results}

\subsection{Main benchmark: the context-switching task}
\label{sec:main_benchmark}

We begin with the main benchmark conditions AB25 and BA30.
These results are reported for representative benchmark conditions rather than for all switch schedules.
Additional switch-time evaluations were also conducted, but they did not support a strong generalization claim and are therefore not emphasized in the main text.
Accordingly, the purpose of the main benchmark is architectural comparison under representative conditions, not a broad claim of switch-time generalization.
A more demanding next step will be to test whether the learned intervention mechanism can respond robustly to randomized or previously unseen switch times, rather than only to representative benchmark schedules.

Figure~\ref{fig:problem_setting} and Figure~\ref{fig:model_overview} summarize the task and the three model families, while Table~\ref{tab:main_performance} reports the main quantitative results.

\begin{figure}[t]
    \centering
    \includegraphics[width=0.9\linewidth]{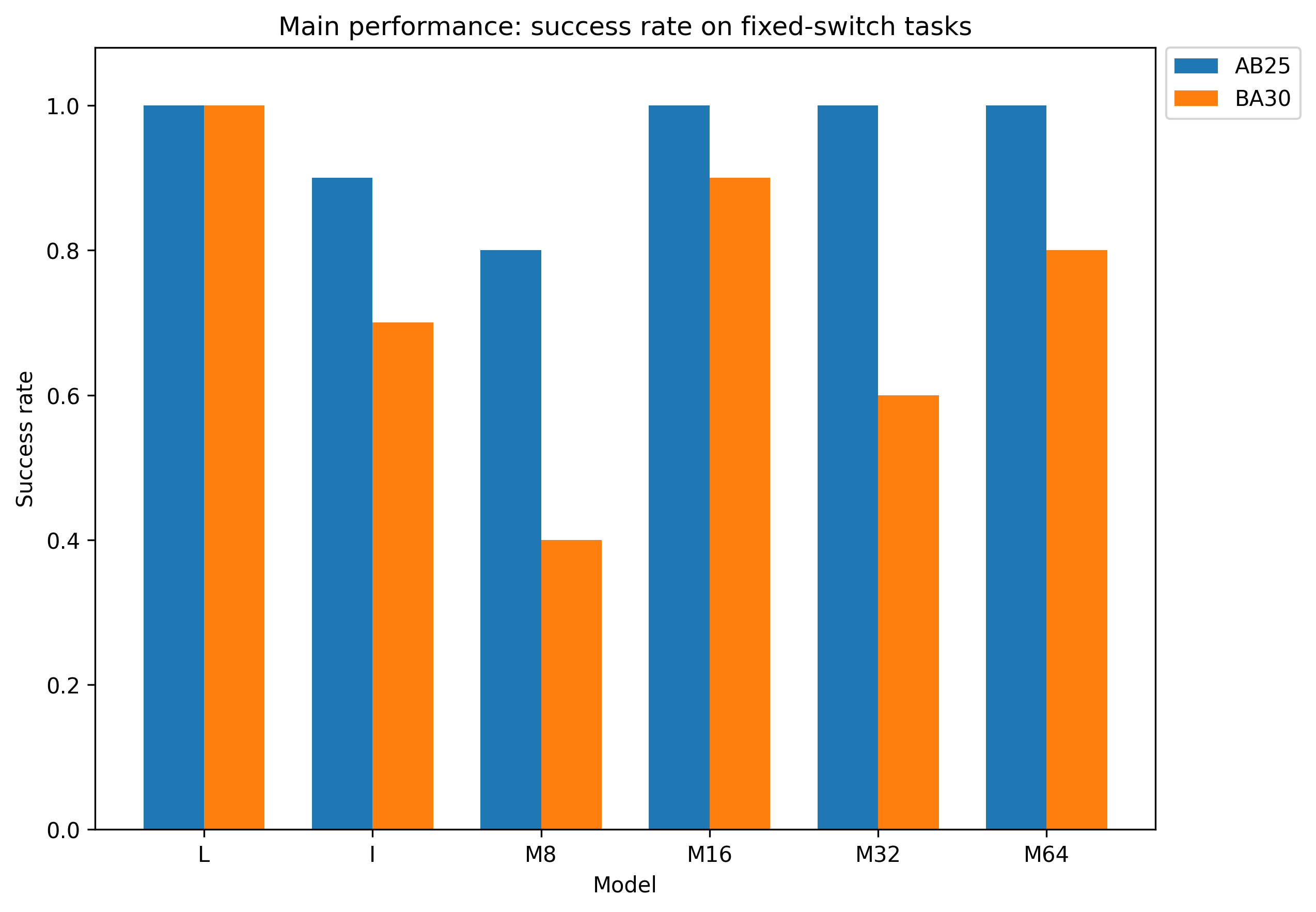}
    \caption{
    Main performance on AB25 and BA30.
    Bars show the fraction of seeds (out of 10) that solved both phases for each model family.
    L solves both tasks perfectly.
    I achieves strong performance without additional recurrent dimensions.
    Among the memory baselines, performance is non-monotonic in memory size: M16 is strongest on BA30 and among the strongest settings on AB25, while both smaller and larger memory settings underperform it on BA30.
    }
    \label{fig:main_performance}
\end{figure}

Figure~\ref{fig:main_performance} reports the main benchmark results.
The label-assisted baseline L achieves perfect performance on both AB25 and BA30.
The proposed intervention model I performs strongly without increasing recurrent dimensionality.
The memory sweep further shows that larger recurrent memory does not lead to monotonic improvement: M16 is the strongest memory-based setting on BA30 and among the strongest on AB25, while M8, M32, and M64 are weaker on BA30.
This pattern supports the view that contextual control is not determined by memory size alone.

\begin{table}[t]
\centering
\caption{Main performance on the context-switching tasks after 300k training steps. ``Success'' reports the number of seeds (out of 10) that solved both phases. ``Phase 1'' reports the mean phase-1 success rate across the same 10 seeds.}
\label{tab:main_performance}
\begin{tabular}{lcccc}
\toprule
Model & AB25 success & BA30 success & AB25 phase 1 & BA30 phase 1 \\
\midrule
L    & 10/10 & 10/10 & 1.00 & 1.00 \\
I    &  9/10 &  7/10 & 0.90 & 0.70 \\
M8   &  8/10 &  4/10 & 0.80 & 0.40 \\
M16  & 10/10 &  9/10 & 1.00 & 0.90 \\
M32  & 10/10 &  6/10 & 1.00 & 0.60 \\
M64  & 10/10 &  8/10 & 1.00 & 0.80 \\
\bottomrule
\end{tabular}
\end{table}

The overall picture is clear.
The label-assisted model L solves both benchmark conditions perfectly, achieving \(10/10\) successful seeds on both AB25 and BA30.
The proposed intervention model I also performs strongly, with \(9/10\) successful seeds on AB25 and \(7/10\) on BA30.
Among the memory baselines, performance depends strongly on the additional memory size:
M8 achieves \(8/10\) on AB25 and \(4/10\) on BA30,
M16 achieves \(10/10\) and \(9/10\),
M32 achieves \(10/10\) and \(6/10\),
and M64 achieves \(10/10\) and \(8/10\).

Two conclusions follow immediately.
First, the intervention model is competitive with the best memory-based baseline while using no additional recurrent dimensions.
Second, increasing memory size is not a monotone route to improvement.
In particular, M16 performs better than both M8 and M32, and M64 recovers only partially relative to M16.
This non-monotonicity is one of the central empirical findings of the paper.

Table~\ref{tab:main_performance} also shows that the main differences across models appear in \textbf{phase 1}.
For L, phase-1 success is \(1.00\) on both tasks.
For I, it is \(0.90\) on AB25 and \(0.70\) on BA30.
For the memory baselines, phase-1 success varies substantially with memory size:
\(0.80/0.40\) for M8,
\(1.00/0.90\) for M16,
\(1.00/0.60\) for M32,
and \(1.00/0.80\) for M64.
Thus, the overall performance ranking is largely explained by how reliably each architecture handles the context-dependent second phase, rather than by failures in the first phase.

The BA30 setting is consistently more difficult than AB25.
This is visible for I as well as for all memory baselines.
We interpret this asymmetry as evidence that the main challenge is not merely reaching goals in the maze, but performing the correct context-dependent reassignment of the target under the more demanding phase-1 condition.
This is precisely the regime in which the architectural differences between L, M, and I become visible.

\subsection{Memory growth is not a sufficient design principle}
\label{sec:memory_growth}

The memory sweep provides a more specific architectural lesson.
If contextual dependence could be handled simply by allocating more recurrent capacity, one would expect performance to improve monotonically with memory size.
Instead, our results show a clear non-monotonic trend.
M16 gives the strongest memory-based performance, while both smaller and larger memory sizes underperform it, especially on BA30.

This observation matters for the conceptual motivation of the paper.
The proposed intervention model is not intended merely as a parameter-saving trick.
Rather, it embodies a different hypothesis about how contextual dependence should be implemented: not by storing more context in an ever larger recurrent state, but by acting on a shared latent state through a context-indexed operator.
The results support this view.
The best memory baseline is strong, but memory growth alone does not provide a uniformly better solution, and the intervention model remains competitive without enlarging the recurrent state at all.

\subsection{Information-theoretic analysis}
\label{sec:info_results}

\begin{table}[t]
\centering
\caption{Estimated conditional contextual information \(I(C;O \mid S)\) (bits) in phase 1 under a uniform prior over contexts. We report mean \(\pm\) standard error across 10 seeds. Goal-related outcomes reveal positive conditional contextual information for all three representative models, including the intervention model without memory growth.}
\label{tab:info_phase1}
\begin{tabular}{llcc}
\toprule
Outcome \(O\) & Model & AB25 & BA30 \\
\midrule
target\_hit & L   & \(0.0372 \pm 0.0125\) & \(0.0370 \pm 0.0078\) \\
target\_hit & I   & \(0.0341 \pm 0.0066\) & \(0.0462 \pm 0.0194\) \\
target\_hit & M16 & \(0.0236 \pm 0.0019\) & \(0.0419 \pm 0.0157\) \\
\addlinespace
goal3 & L   & \(0.0410 \pm 0.0143\) & \(0.0427 \pm 0.0104\) \\
goal3 & I   & \(0.0402 \pm 0.0097\) & \(0.0550 \pm 0.0240\) \\
goal3 & M16 & \(0.0254 \pm 0.0025\) & \(0.0440 \pm 0.0170\) \\
\bottomrule
\end{tabular}
\end{table}

We next turn to the empirical quantity \(I(C;O\mid S)\), which we use as an operational probe of contextual dependence at fixed latent state.
As discussed in Section~\ref{sec:info_quantities}, the outcome definition \(O\) is crucial.
In the main text we therefore focus on task-relevant phase-1 outcomes, reported in Table~\ref{tab:info_phase1}.

For the \texttt{target\_hit} outcome, all three representative models exhibit positive conditional contextual information under a uniform context prior.
On AB25, the estimated values are \(0.0372 \pm 0.0125\) bits for L, \(0.0341 \pm 0.0066\) bits for I, and \(0.0236 \pm 0.0019\) bits for M16.
On BA30, the corresponding values are \(0.0370 \pm 0.0078\), \(0.0462 \pm 0.0194\), and \(0.0419 \pm 0.0157\) bits.

For the coarser \texttt{goal3} outcome, which distinguishes \(\{\text{other}, \text{target}, \text{wrong}\}\), the same pattern remains.
On AB25, the estimates are \(0.0410 \pm 0.0143\) bits for L, \(0.0402 \pm 0.0097\) bits for I, and \(0.0254 \pm 0.0025\) bits for M16.
On BA30, they are \(0.0427 \pm 0.0104\), \(0.0550 \pm 0.0240\), and \(0.0440 \pm 0.0170\), respectively.

These results establish two important points.
First, the proposed intervention model exhibits positive conditional contextual information for task-relevant outcomes in phase 1, consistent with its intended role as a mechanism for context-dependent control without memory growth.
Second, the relevant contextual effect is more naturally expressed at the level of goal-related outcomes than at the level of immediate primitive actions.
This supports our decision to operationalize the theorem-motivated probe using task-level outcomes rather than insisting on a one-step action-level definition.

At the same time, the information-theoretic analysis should be interpreted with appropriate care.
Our goal is not to claim a complete numerical verification of every assumption or term in the theorem of \cite{Kim2026NoGo}.
Rather, the results show that once \(O\) is defined in a manner aligned with the task semantics of the context-switching task, the intervention model carries measurable positive contextual information at fixed latent state.
In particular, the positivity of \(I(C;O\mid S)\) shows that contextual dependence is not fully screened off by the latent state \(S\) used in our estimator.
We do not interpret the observed values as large in an absolute sense.
A likely reason they remain numerically small is that the estimator averages over many latent states in which behavior is dominated by generic navigation, whereas context-dependent differences matter most at a relatively small subset of task-relevant phase-1 decision points.
This is also consistent with the structure of the benchmark, in which only a subset of phase-1 states lie near context-sensitive goal-selection points, while many other states are dominated by context-independent navigation.

\section{Discussion}
\label{sec:discussion}

The main contribution of this paper is architectural.
We proposed an intervention-based recurrent implementation of contextual dependence and showed that it works competitively on the context-switching task without enlarging recurrent dimensionality.
This should not be read as showing that intervention universally dominates either explicit context input or memory expansion.
On this benchmark, the label-assisted model L remains a strong oracle-style reference baseline, which is expected because the contextual variable is directly available at every step.
Likewise, a moderate amount of additional recurrent memory already solves much of the task, as illustrated by the strong performance of the best memory baseline.
The main point is therefore more specific: the intervention model realizes contextual dependence through a distinct mechanism, namely a context-indexed transformation of a shared pre-intervention latent state.

The relation to the information-theoretic theorem also requires a precise interpretation.
Kim \cite{Kim2026NoGo} proves that, for classical single-state ontological models with fixed ontic-state reuse across interventions, contextual dependence requires an auxiliary contextual variable satisfying
\[
H(M_{\mathrm{aux}})\ge I(C;O\mid \lambda) > 0.
\]
Here again, \(M_{\mathrm{aux}}\) denotes the theorem's auxiliary contextual variable and should not be confused with our memory baseline \(M\).
The two play different roles:
\(M\) is an empirical model family with enlarged recurrent hidden state, whereas \(M_{\mathrm{aux}}\) is a theorem-level resource required in a classical single-state representation.
Our experiments do not instantiate that theorem literally.
The conditioning variable in our estimator is a learned recurrent latent state, not an ontic variable \(\lambda\), and we do not directly estimate the minimal auxiliary variable appearing in the theorem.
What we do show is narrower but still meaningful: in an operational analogue motivated by the theorem, the intervention model exhibits positive \(I(C;O\mid S)\) for task-relevant phase-1 outcomes.
In this sense, the theorem serves primarily as a conceptual resource-accounting framework for our study, rather than as a mathematically exact description of the learned recurrent dynamics.

The positivity of \(I(C;O\mid S)\) has a precise interpretation in our setting.
It shows that, even after conditioning on the latent state used in our estimator, the observable outcome distribution still depends on context.
In other words, contextual dependence is not fully screened off by the conditioned latent state alone.
This is the precise sense in which our results are consistent with the theorem's resource-accounting picture: they provide empirical evidence that conditioning on a shared latent representation does not by itself eliminate context dependence at the level of observable outcomes.
At the same time, this does not identify the theorem's ontic variable \(\lambda\), nor does it directly estimate the minimal auxiliary contextual variable \(M_{\mathrm{aux}}\).
For L and I, the explicit contextual signal used by the architecture is binary, giving a simple architecture-level upper bound of at most one bit under a uniform context prior. We treat this only as an architecture-level interpretation for those two models, not as a direct estimate of the theorem's \(M_{\mathrm{aux}}\).
Thus, we interpret the present results as support for the theorem's qualitative implication, rather than as a complete numerical verification of the theorem or its minimality claim.
The point is therefore structural rather than magnitude-based: the relevant conclusion is that context still changes the task-relevant outcome distribution after conditioning on \(S\), not that the measured value is numerically close to its theoretical upper bound.

A further important observation is that raw capacity increase is not, by itself, a complete design principle for contextual control.
Within the memory-baseline family, performance does not improve monotonically with added recurrent dimensions.
We do not interpret this as a statement about the theorem's \(M_{\mathrm{aux}}\), and it should not be read that way.
Rather, it is an architectural observation about one particular baseline family: simply enlarging recurrent hidden state does not guarantee better contextual control.
This strengthens the motivation for studying intervention as a distinct implementation principle rather than treating contextual dependence only as a hidden-size scaling problem.
One possible interpretation is that simple memory growth leaves the model to discover both state representation and contextual routing implicitly within the same recurrent dynamics.
By contrast, the intervention architecture factorizes these roles: the recurrent core builds a shared latent representation, while context acts through an explicit, structured modulation of that representation.
We do not claim that this factorization is universally superior, but it offers a plausible explanation for why contextual control can remain strong without increasing recurrent dimensionality.

An additional reason this factorization may matter is that the intervention used here is a residual linear map acting on a shared latent state. In geometric terms, such a map can be understood as a context-dependent local transformation of the latent representation, rather than as a demand that the recurrent core itself separately encode and route all contextual variation. In this sense, the intervention can be viewed as shifting or reorienting the task-relevant latent geometry in a context-dependent way, while leaving the recurrent core responsible for building the shared representation itself. We do not claim that linear intervention is universally sufficient, but in the present benchmark it provides a minimal and interpretable mechanism for shifting task-relevant latent structure without increasing recurrent dimensionality.

Although the present benchmark is deliberately small and controlled, the underlying architectural issue is relevant to larger agentic systems, in which contextual dependence often appears as task-phase switching, role switching, or tool-selection control over a shared internal state.
At the same time, there are clear limitations.
First, the experiments are conducted on a controlled gridworld benchmark rather than on large-scale partially observable domains.
Second, our information-theoretic analysis is tied to specific task-relevant outcome definitions and a particular counterfactual estimator.
Third, we do not claim that explicit context input is already too costly in the present setting; with a binary context variable and a compact benchmark, direct conditioning is simple and effective.
An additional limitation is that the intervention used here is deliberately simple: \(D_c\) is a learned linear map. We chose this form as a minimal and interpretable test of context-dependent control over a shared latent state, rather than as the most expressive possible conditional mechanism.
We also fixed the intervention strength \(\alpha\) in the present study, rather than learning it jointly with the rest of the model. In limited preliminary tuning, we tested \(\alpha=0.1, 0.2,\) and \(0.3\), and found \(\alpha=0.1\) to give the most satisfactory overall behavior in the present setup.
This choice keeps the intervention as a controlled residual perturbation and preserves the near-identity initialization induced by the zero-initialized operators at the start of training. At the same time, we do not interpret this as a complete sensitivity analysis: the preferred value of \(\alpha\) may depend on other hyperparameter choices, and a more systematic study of the trade-off between intervention strength, optimization stability, and final performance remains future work.
We also did not include direct empirical comparisons against other conditional-modulation mechanisms such as FiLM-style conditioning or hypernetwork-based modulation. Such comparisons would help clarify which aspects of the present results are specific to the simple residual linear intervention used here.

We also conducted preliminary checks at unseen switch times, but the resulting generalization was partial and asymmetric across orders, so we do not treat switch-time generalization as a main claim here.
At the same time, because the intervention acts directly on the shared latent state after the recurrent update, the architecture has a structurally modular form that may permit more immediate reactions to context changes than approaches that must encode contextual routing only implicitly within recurrent memory.
A particularly important direction for future work is to evaluate whether the intervention mechanism can support zero-shot or near-zero-shot adaptation to randomized switch schedules, where context changes occur at times not seen during training.
These limitations nonetheless point naturally to future work.
One direction is to extend intervention-based recurrent architectures to richer partially observable control tasks with more complex context structure.
Another is to study whether the same design principle can be combined with stronger recurrent backbones or transformer-style sequence models.
A third is to develop tighter empirical estimators of the information-theoretic quantities involved in theorem-motivated analyses of contextual control.
A useful next step would also be to visualize, for fixed latent states, how the intervention changes the policy distribution or goal-level outcome distribution under counterfactual context changes.
Related analyses of latent geometry, such as cosine similarity or low-dimensional projections of pre- and post-intervention states, would also help clarify how the linear intervention reshapes task-relevant latent structure.
In that sense, the intervention-based design studied here may be useful not only for compact recurrent control problems but also as a lightweight context-routing mechanism in larger agentic architectures.

\section{Conclusion}
\label{sec:conclusion}

We introduced an intervention-based recurrent architecture for contextual control and evaluated it on the context-switching benchmark.
The proposed model realizes contextual dependence by applying a context-indexed operator to a shared recurrent latent state, thereby avoiding recurrent memory growth.

Empirically, the intervention model performs strongly on the main benchmark and is competitive with the best memory-based baseline while using no additional recurrent dimensions.
At the same time, the memory sweep shows that increasing recurrent capacity is not a monotone route to better contextual control.
This highlights the importance of architectural mechanism, not just model size.

Our information-theoretic analysis further shows that the intervention model exhibits positive \(I(C;O\mid S)\) for task-relevant goal-level outcomes in phase 1.
We interpret this not as a direct theorem verification, but as an operational probe motivated by recent single-state information-theoretic analyses of contextuality.
Together with the simple architecture-level upper-bound interpretation for explicit binary contextual signals, this provides a principled connection between the architecture and the motivating theoretical picture.

Overall, our results support the view that contextual dependence need not be implemented by ever larger recurrent memory.
Intervention on a shared latent state offers a simple and effective alternative in this setting, and may also provide a useful architectural primitive for context-dependent control in larger agentic systems.

\section*{Data Availability Statement}
The code, evaluation scripts, and aggregated result files necessary to reproduce the main tables and figures are publicly available at \url{https://github.com/songju1/Contextual-Control}.
Additional intermediate logs and auxiliary files are available from the corresponding author upon reasonable request.

\section*{Acknowledgments}

This work was supported by SOBIN Institute LLC under Research Grant SP008.  
The authors used ChatGPT (OpenAI) to improve the English language and grammatical correctness of the manuscript.
After using this tool, the authors reviewed and edited the content as needed and take full responsibility for the final version of the manuscript.

\bibliographystyle{plain}
\bibliography{refs}

@article{KaelblingLittmanCassandra1998,
  author  = {Leslie Pack Kaelbling and Michael L. Littman and Anthony R. Cassandra},
  title   = {Planning and Acting in Partially Observable Stochastic Domains},
  journal = {Artificial Intelligence},
  volume  = {101},
  number  = {1--2},
  pages   = {99--134},
  year    = {1998}
}

@article{HochreiterSchmidhuber1997,
  author  = {Sepp Hochreiter and J{\"u}rgen Schmidhuber},
  title   = {Long Short-Term Memory},
  journal = {Neural Computation},
  volume  = {9},
  number  = {8},
  pages   = {1735--1780},
  year    = {1997},
  doi     = {10.1162/neco.1997.9.8.1735}
}

@misc{HausknechtStone2015,
  author        = {Matthew Hausknecht and Peter Stone},
  title         = {Deep Recurrent Q-Learning for Partially Observable {MDP}s},
  year          = {2015},
  eprint        = {1507.06527},
  archivePrefix = {arXiv},
  primaryClass  = {cs.LG},
  note          = {arXiv:1507.06527 [cs.LG]}
}

@inproceedings{PerezEtAl2018,
  author    = {Ethan Perez and Florian Strub and Harm de Vries and Vincent Dumoulin and Aaron Courville},
  title     = {{FiLM}: Visual Reasoning with a General Conditioning Layer},
  booktitle = {Proceedings of the AAAI Conference on Artificial Intelligence},
  volume    = {32},
  number    = {1},
  year      = {2018}
}

@article{Spekkens2005,
  author  = {Robert W. Spekkens},
  title   = {Contextuality for Preparations, Transformations, and Unsharp Measurements},
  journal = {Physical Review A},
  volume  = {71},
  pages   = {052108},
  year    = {2005},
  doi     = {10.1103/PhysRevA.71.052108}
}

@article{AbramskyBrandenburger2011,
  author  = {Samson Abramsky and Adam Brandenburger},
  title   = {The Sheaf-Theoretic Structure of Non-Locality and Contextuality},
  journal = {New Journal of Physics},
  volume  = {13},
  pages   = {113036},
  year    = {2011},
  doi     = {10.1088/1367-2630/13/11/113036}
}

@misc{Kim2026DecisionDynamics,
  author        = {Song-Ju Kim},
  title         = {Contextuality Derived from Minimal Decision Dynamics: Quantum Tug-of-War Decision Making},
  year          = {2026},
  eprint        = {2601.10034},
  archivePrefix = {arXiv},
  primaryClass  = {quant-ph},
  note          = {arXiv:2601.10034 [quant-ph]}
}

@misc{Kim2026Obstruction,
  author        = {Song-Ju Kim},
  title         = {Contextuality as an Information-Theoretic Obstruction to Classical Probability},
  year          = {2026},
  eprint        = {2601.20167},
  archivePrefix = {arXiv},
  primaryClass  = {quant-ph},
  note          = {arXiv:2601.20167 [quant-ph]}
}

@misc{Kim2026NoGo,
  author        = {Song-Ju Kim},
  title         = {Contextuality from Single-State Ontological Models: An Information-Theoretic No-Go Theorem},
  year          = {2026},
  eprint        = {2602.16716},
  archivePrefix = {arXiv},
  primaryClass  = {quant-ph},
  note          = {arXiv:2602.16716 [cs.AI] [quant-ph]}
}

@inproceedings{Bakker2001RLLSTM,
  author    = {Bram Bakker},
  title     = {Reinforcement Learning with Long Short-Term Memory},
  booktitle = {Advances in Neural Information Processing Systems 14},
  year      = {2001}
}

@inproceedings{KapturowskiEtAl2019R2D2,
  author    = {Steven Kapturowski and Georg Ostrovski and John Quan and R{\'e}mi Munos and Will Dabney},
  title     = {Recurrent Experience Replay in Distributed Reinforcement Learning},
  booktitle = {International Conference on Learning Representations},
  year      = {2019}
}

@inproceedings{IglEtAl2018DVRL,
  author    = {Maximilian Igl and Luisa Zintgraf and Tuan Anh Le and Frank Wood and Shimon Whiteson},
  title     = {Deep Variational Reinforcement Learning for {POMDP}s},
  booktitle = {Proceedings of the 35th International Conference on Machine Learning},
  series    = {Proceedings of Machine Learning Research},
  volume    = {80},
  pages     = {2117--2126},
  year      = {2018}
}

@inproceedings{LittmanSuttonSingh2001PSR,
  author    = {Michael L. Littman and Richard S. Sutton and Satinder Singh},
  title     = {Predictive Representations of State},
  booktitle = {Advances in Neural Information Processing Systems 14},
  pages     = {1555--1561},
  year      = {2001}
}

@inproceedings{deVriesEtAl2017CBN,
  author    = {Harm de Vries and Florian Strub and J{\'e}r{\'e}mie Mary and Hugo Larochelle and Olivier Pietquin and Aaron Courville},
  title     = {Modulating Early Visual Processing by Language},
  booktitle = {Advances in Neural Information Processing Systems 30},
  year      = {2017}
}

@article{DumoulinEtAl2018FeatureWise,
  author  = {Vincent Dumoulin and Ethan Perez and Nathan Schucher and Florian Strub and Harm de Vries and Aaron Courville and Yoshua Bengio},
  title   = {Feature-wise Transformations},
  journal = {Distill},
  volume  = {3},
  number  = {7},
  pages   = {e11},
  year    = {2018},
  doi     = {10.23915/distill.00011}
}

@inproceedings{KimSongBengio2017DLN,
  author    = {Taesup Kim and Inchul Song and Yoshua Bengio},
  title     = {Dynamic Layer Normalization for Adaptive Neural Acoustic Modeling in Speech Recognition},
  booktitle = {Proceedings of Interspeech 2017},
  pages     = {3317--3321},
  year      = {2017},
  doi       = {10.21437/Interspeech.2017-1262}
}

\end{document}